\documentclass[10pt,twocolumn,letterpaper]{article}

\usepackage[pagenumbers]{cvpr} 

\usepackage{amsmath,amssymb}
\usepackage{algorithm}
\usepackage{algpseudocode}

\usepackage{graphicx}

\usepackage{multirow}
\usepackage[table]{xcolor}
\usepackage{wasysym}
\usepackage{bm}
\usepackage{booktabs}
\usepackage{tabularx}
\usepackage{colortbl}
\usepackage{microtype}
\usepackage{makecell}
\usepackage{pifont}

\newcommand{\cmark}{\ding{51}} 
\newcommand{\xmark}{\ding{55}} 

\makeatletter
\@ifundefined{NC@find@C}{\newcolumntype{C}{>{\centering\arraybackslash}X}}{}
\makeatother

\definecolor{cvprblue}{rgb}{0.21,0.49,0.74}
\usepackage[pagebackref,breaklinks,colorlinks,allcolors=cvprblue]{hyperref}

\def\confName{CVPR}
\def\confYear{2027}

\title{Beyond Feature Fusion: Sensor-Grounded Semantic–Spiking Learning for Robust Object Detection}

\author{Hanwen~Liang$^{1}$ \quad Wenxuan~Liu$^{2}$ \quad Wenxin~Huang$^3$ \quad Jingling~Yuan$^1$ \quad Zhaofei~Yu$^2$\\ 
\quad Xian~Zhong$^{1}$  \quad Tiejun~Huang$^2$ \\
$^1$ Wuhan University of Technology \quad 
$^2$ Peking University \quad
$^3$ Hubei University \\
}

\begin{document}
\maketitle

\begin{abstract}
Underexposure weakens object details, and overexposure obscures boundaries. Spike cameras encode light into high-frequency binary streams. Long-window spike counts give stable appearance cues, while short observations reveal changes in firing rates from sparse measurements. We propose \textbf{SPIRE}, a sensor-grounded semantic-spiking framework for illumination-robust object detection from a single spike stream. SPIRE builds an appearance view from the full packet through detection-oriented illumination adaptation and constructs a sequence of cumulative views from progressively longer prefixes. Each prefix supplies a rate map, an edge map, and a rate difference that tracks changes as spikes accumulate. At selected scales, ANN features and spike-derived initialization jointly set the initial membrane potentials and firing thresholds, allowing semantic cues to shape the responses during temporal encoding. Local spike-count contrasts supervise both branches during training. The residual between measured and ANN-predicted contrasts combines with the spiking features to produce a bounded correction to the fused feature pyramid. On simulated VOC and COCO, SPIRE-D outperforms the strongest baseline by 4.00-5.50 AP$_{50}$ points across the four under- and overexposure settings. On real PKU-Vidar-DVS recordings, it achieves 67.07 AP$_{50}$ and 40.10 AP$_{75}$. SPIRE-D also uses 50.4\% fewer MACs than SPFD.
\end{abstract}

\section{Introduction}

\begin{figure}[!t]
    \centering
    \includegraphics[width=\linewidth]{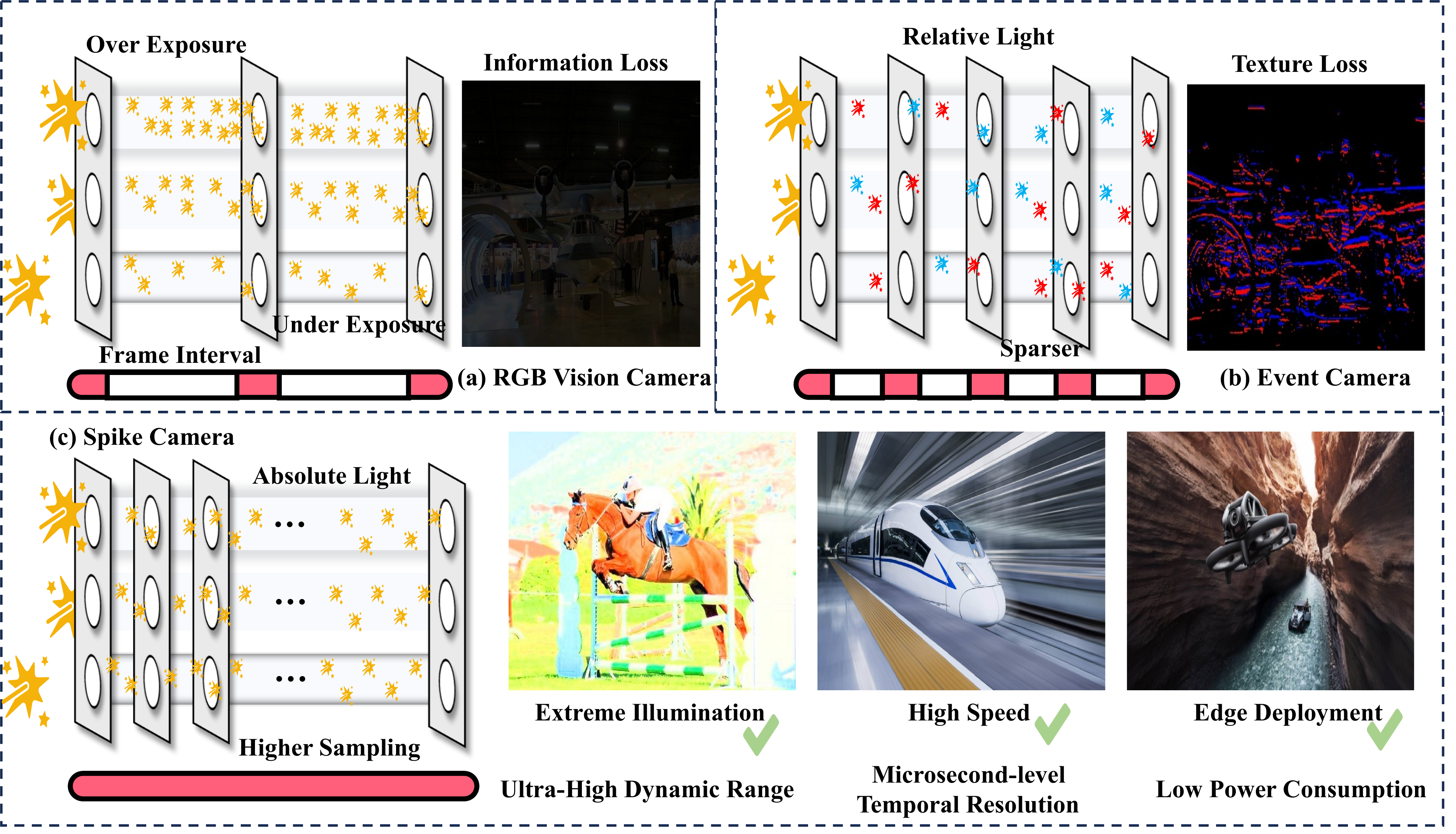}
    \caption{\textbf{Imaging comparison.} RGB cameras suffer from exposure-dependent degradation, while event cameras capture relative brightness changes. Spike cameras retain intensity-related information through high-frequency binary sampling.}
    \label{fig:sensing_motivation}
\end{figure}

Extreme illumination remains a major challenge for object detection in autonomous driving, robotics, and surveillance. Underexposure weakens visual details and amplifies noise, while overexposure obscures discriminative structures. Although task-oriented image processing and RAW-domain detection alleviate such degradation~\cite{liu2022iayolo, AODRaw_cvpr2025,guo2025darkisp}, frame-based cameras remain constrained by exposure-dependent information acquisition. As illustrated in \cref{fig:sensing_motivation}, event cameras provide high temporal resolution but primarily capture relative brightness changes, offering limited intensity information in static regions~\cite{gallego2022survey,liu2024night}. In contrast, spike cameras encode accumulated incident light through high-frequency binary measurements~\cite{huang2023camera,zhu2023fovea}, providing a promising sensing modality for illumination-robust object detection.

A spike packet supports several complementary measurements. Integrating over the full packet stabilizes the appearance estimate, as exploited in spike reconstruction and perception~\cite{zhao2021spk2img,dong2024bayer,liu2025easd}. Observing the same packet at successive times exposes changes in firing rates that a single average compresses. Spatial differences in accumulated counts capture local structure, including in stationary regions. These signals serve different roles in detection. The full packet supports semantic recognition, the sequence of observations drives temporal encoding, and spatial count contrasts provide a reference for refining the fused features. RGB-event and hybrid ANN-SNN detectors demonstrate the value of combining complementary features~\cite{cao2024eolo,ahmed2025hybrid,wang2026spfd}. SPIRE connects the two representations during neuronal encoding and brings the count measurements back into the detection path after fusion.

To this end, we propose \textbf{SPIRE}, a sensor-grounded semantic-spiking illumination-robust estimation for object detection using a single spike modality. To reconcile stable appearance estimation with temporal evidence preservation, SPIRE constructs two complementary views from the same spike stream. A semantic appearance branch employs detection-oriented illumination adaptation and an ANN encoder to extract discriminative semantic features from full-window spike statistics. In parallel, a temporal-structural branch constructs cumulative rate, edge, and inter-prefix difference representations, which are processed by a spiking encoder. These cumulative observations capture the evolution of spike statistics while mitigating the instability of isolated short-window measurements.

SPIRE introduces \textit{semantic-guided spiking encoding} to connect the two branches during feature generation. Inspired by neuronal configuration mechanisms~\cite{lin2025clearsight}, we combine ANN-derived semantic prompts with spike-derived initialization to configure the initial membrane potentials and firing thresholds of spiking neurons at selected scales. These initial states and thresholds let semantic cues shape the spiking responses before cross-stream feature fusion.

Furthermore, we introduce \textit{measurement-guided refinement} to maintain an explicit connection between learned representations and raw spike observations. Specifically, normalized contrasts between neighboring spike counts provide observation-derived structural references for supervising both semantic and spiking branches. During inference, SPIRE combines the ANN contrast residual with the spiking features to produce a bounded correction to the fused features. We evaluate SPIRE under simulated under- and overexposure and on real spike recordings.

In summary, our main contributions are threefold.
\begin{itemize}
    \item We establish a single-source spike detection framework that jointly exploits complementary semantic appearance and cumulative temporal-structural representations.
    \item We develop semantic-guided spiking encoding that incorporates ANN semantics and spike-derived initialization into neuronal state configuration, enabling interaction during temporal encoding.
    \item We introduce measurement-guided refinement that leverages raw spike-count contrasts for cross-branch supervision and bounded feature correction, achieving robust detection across challenging illumination conditions.
\end{itemize}

\section{Related Work}

\subsection{Detection under Challenging Illumination}
Illumination-robust detection has been explored through task-aware image processing~\cite{liu2022iayolo}, day-night adaptation~\cite{du2024daynight}, and RAW-domain learning~\cite{AODRaw_cvpr2025,guo2025darkisp}. Event cameras further support challenging-light perception, motivating nighttime reconstruction~\cite{liu2024night} and RGB-event detection~\cite{cao2024eolo,wang2026spfd}. In particular, SPFD~\cite{wang2026spfd} disentangles shared and modality-specific features to exploit complementary RGB-event information. SPIRE studies the single-sensor setting, where appearance and temporal structure must both be recovered from an intensity-encoding spike stream.

\subsection{Spike-Camera Representation and Perception}
Spike cameras encode intensity-related information through high-frequency binary measurements~\cite{huang2023camera,zhu2023fovea}. Prior studies investigate image reconstruction~\cite{zhao2021spk2img,zheng2021stp}, optical flow~\cite{zhao2022flow}, color recovery, and temporal modeling~\cite{dong2024sjdd,dong2024bayer,wang2026spk2vid}. For object detection, Driving in Spikes~\cite{liu2025easd} exploits complementary spike representations and entropy-guided feature selection for object detection. SPIRE uses cumulative views for temporal encoding, configures the spiking neurons with semantic cues, and refines detection features with raw spike-count contrasts.

\subsection{Event Detection and Semantic-Spiking Interaction}
Event-based detection has advanced through recurrent attention~\cite{gehrig2023rvt}, sparse Transformers~\cite{peng2024sast}, and adaptive spiking representations~\cite{wang2024eassnn,li2026sda,li2025msd}. Related methods explore semantic feature refinement~\cite{li2025seenda} and hybrid ANN-SNN perception~\cite{ahmed2025hybrid}, primarily emphasizing feature transformation and interaction. ClearSight~\cite{lin2025clearsight} further introduces semantic-guided neuronal configuration for event-assisted deblurring. SPIRE applies semantic-guided neuronal configuration to cumulative spike representations for detection. It also uses intensity-related information in stationary regions to guide feature refinement with raw spike counts.

\section{Proposed Method}
\label{sec:method}

\subsection{Overall Architecture}

\begin{figure*}[t]
    \centering
    \includegraphics[width=\textwidth]{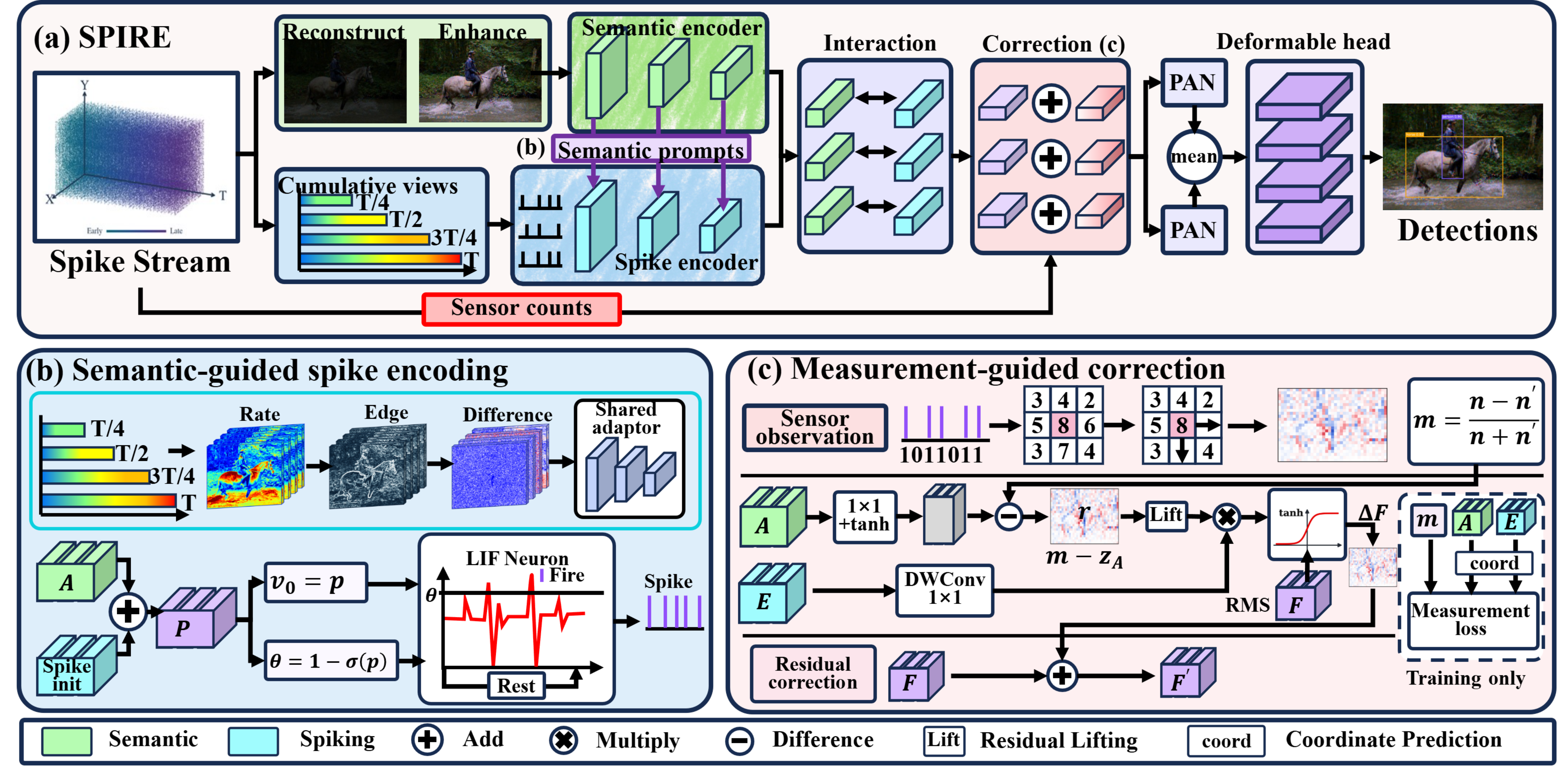}
    \caption{\textbf{Overview of SPIRE.} (a) A single spike packet provides a full-window appearance view for the ANN and a sequence of cumulative views for the SNN. The detector receives the fused and refined feature pyramid. (b) ANN prompts and spike-derived initialization set the initial membrane potentials and firing thresholds before the four SNN steps. (c) The dashed region marks training-time supervision of both contrast heads. The ANN contrast residual and the SNN readout drive the feature correction during training and inference.}
\label{fig:spire_overview}
\end{figure*}

Given a spike packet $\left\{\bm S_t\right\}_{t=1}^{T}$, with $\bm S_t\in\left\{0,1\right\}^{B\times C_s\times H\times W}$, SPIRE predicts object categories and bounding boxes using a single spike modality. Here, $B$ is the batch size and $C_s=1$ for monochrome measurements or $C_s=4$ for compact RGGB measurements. As shown in \cref{fig:spire_overview}(a), SPIRE constructs two complementary representations from the same packet: an appearance view for ANN-based semantic encoding and cumulative structural views for SNN-based temporal encoding. Semantic cues further configure spiking neurons at selected stages. Multi-scale features are then aligned and fused, followed by a correction mechanism anchored to the original spike counts. The refined features pass through two parallel path-aggregation networks (PANs) and a Deformable DETR detection head.

\subsection{Complementary Spike Representations}

\paragraph{Semantic appearance view.}
Accumulating spikes across the full packet provides an intensity-related estimate suitable for a pretrained visual encoder. We compute the firing-rate map $\bar{\bm S}=T^{-1}\sum_{t=1}^{T}\bm S_t$ and construct an appearance input:
\begin{equation}
    \begin{aligned}
        \bm I &= \operatorname{clip} \left(\mathcal R \left(\kappa\bar{\bm S}\right),0,1\right),\\
        \bm X^a &= \mathcal I_{\mathrm{ISP}} \left(\bm I,\bm Q\right),
    \end{aligned}
\end{equation}
where $\mathcal R$ denotes local CFA interpolation for RGGB measurements or channel replication for monochrome input, and $\kappa$ is a fixed rate-to-intensity factor. The spike-derived cues $\bm Q$ summarize rate, recent firing intervals, temporal variation, stability, entropy, and confidence. A lightweight, differentiable ISP predicts bounded white-balance, exposure, gamma, contrast, and sharpening adjustments conditioned on $\bm I$ and $\bm Q$. Following detection-oriented image processing~\cite{liu2022iayolo}, we optimize this frontend with the detection loss. The ANN encoder yields multi-scale features $\bm H_l=\left[\mathcal B_a \left(\bm X^a\right)\right]_l$, which are projected, when needed, to fusion features $\bm A_l=\Pi_l \left(\bm H_l\right)$.

\paragraph{Cumulative structural view.}
A full-window rate map omits how firing statistics evolve within the packet. We use $K=4$ cumulative prefixes with endpoints $t_k=\operatorname{round} \left(kT/K\right)$. Each prefix is converted to a grayscale rate, a normalized Scharr edge map, and an inter-prefix difference:
\begin{equation}
    \begin{aligned}
        \bm r_k &= \operatorname{Gray} \left(\frac{1}{t_k}\sum_{t=1}^{t_k}\bm S_t\right),\\
        \bm e_k &= \operatorname{Norm} \left(\left\lVert\nabla_{\mathrm{Scharr}}\bm r_k\right\rVert_2\right),\\
        \bm d_k &=
        \begin{cases}
            \bm r_1, & k=1,\\
            \left\lvert\bm r_k-\bm r_{k-1}\right\rvert, & k>1,
        \end{cases}\\
        \bm X_k &= \mathcal A \left(\left[\bm r_k,\bm e_k,\bm d_k\right]\right),
    \end{aligned}
\end{equation}
where $\operatorname{Gray}$ includes CFA interpolation when applicable and $\operatorname{Norm}$ applies per-map maximum normalization to the pointwise Scharr magnitude. The brackets $\left[\cdot\right]$ denote channel concatenation. The shared residual adapter $\mathcal A$ maps each descriptor to a three-channel SNN input. As illustrated in \cref{fig:spire_overview}(b), the four views share a starting point and cover progressively longer prefixes. For a 200-frame packet, the prefix endpoints are 50, 100, 150, and 200 frames. Each rate is normalized by its own prefix length before computing the difference map. The rate and edge channels describe the structure accumulated up to each endpoint, while the difference channel indicates where the estimated rate has changed since the previous endpoint.

\subsection{Semantic-Guided Spiking Encoding}

As shown in \cref{fig:spire_overview}(b), the four adapted inputs are processed in order by a Spike-CSP encoder. Following ClearSight~\cite{lin2025clearsight}, we combine an ANN prompt with a spike-derived spatial initialization map:
\begin{equation}
    \begin{aligned}
        \bm q &= \mathcal P_s \left(\frac{1}{K}\sum_{k=1}^{K}\bm X_k\right),\\
        \bm p_l &= \mathcal P_l \left(\bm H_l\right)
           +\operatorname{Resize}_l \left(\bm q\right),\quad l\in\left\{3,4\right\},\\
        \bm v_{l,0} &= \bm p_l,\qquad
        \boldsymbol\theta_l=1-\sigma \left(\bm p_l\right),
    \end{aligned}
\end{equation}
where $\mathcal P_s$ is a $1\times1$ projection to a single-channel map, $\mathcal P_l$ projects the ANN features to the native channels of SNN stage $l$, and $\sigma$ denotes the sigmoid function. The resized spike initialization is broadcast over channels. The prompt is computed once per packet at each configured stage and sets a threshold that is shared across the four steps. At a given location and channel, a larger value in $\bm p_l$ raises the starting potential and lowers the threshold, making an initial spike easier to elicit for the same input current. Subsequent firing depends on the accumulated state and the reset in \cref{eq:lif_dynamics}.

Given the input current $\bm J_{l,k}$ at SNN step $k$, the configured leaky integrate-and-fire neuron follows:
\begin{equation}
    \label{eq:lif_dynamics}
    \begin{aligned}
        \widetilde{\bm v}_{l,k}
          &=\bm v_{l,k-1}
            +\frac{\bm J_{l,k}-\bm v_{l,k-1}}{\tau},\\
        \bm o_{l,k}
          &=\mathbb I \left[\widetilde{\bm v}_{l,k}\geq\boldsymbol\theta_l\right],\\
        \bm v_{l,k}
          &=\widetilde{\bm v}_{l,k}
            -\boldsymbol\theta_l\odot\bm o_{l,k},
    \end{aligned}
\end{equation}
where $\tau=2$, $\mathbb I \left[\cdot\right]$ is the firing function, and $\odot$ denotes element-wise multiplication throughout. Surrogate gradients enable optimization. The configured stages additionally include voltage-initialized alignment and residual projection; P5 uses ordinary spiking processing. We denote the resulting multi-scale sequences by $\bm U_l$. Neuronal states evolve across the four distinct cumulative inputs and are reset between packets.

\begin{figure}[t]
    \centering
    \includegraphics[width=\linewidth]{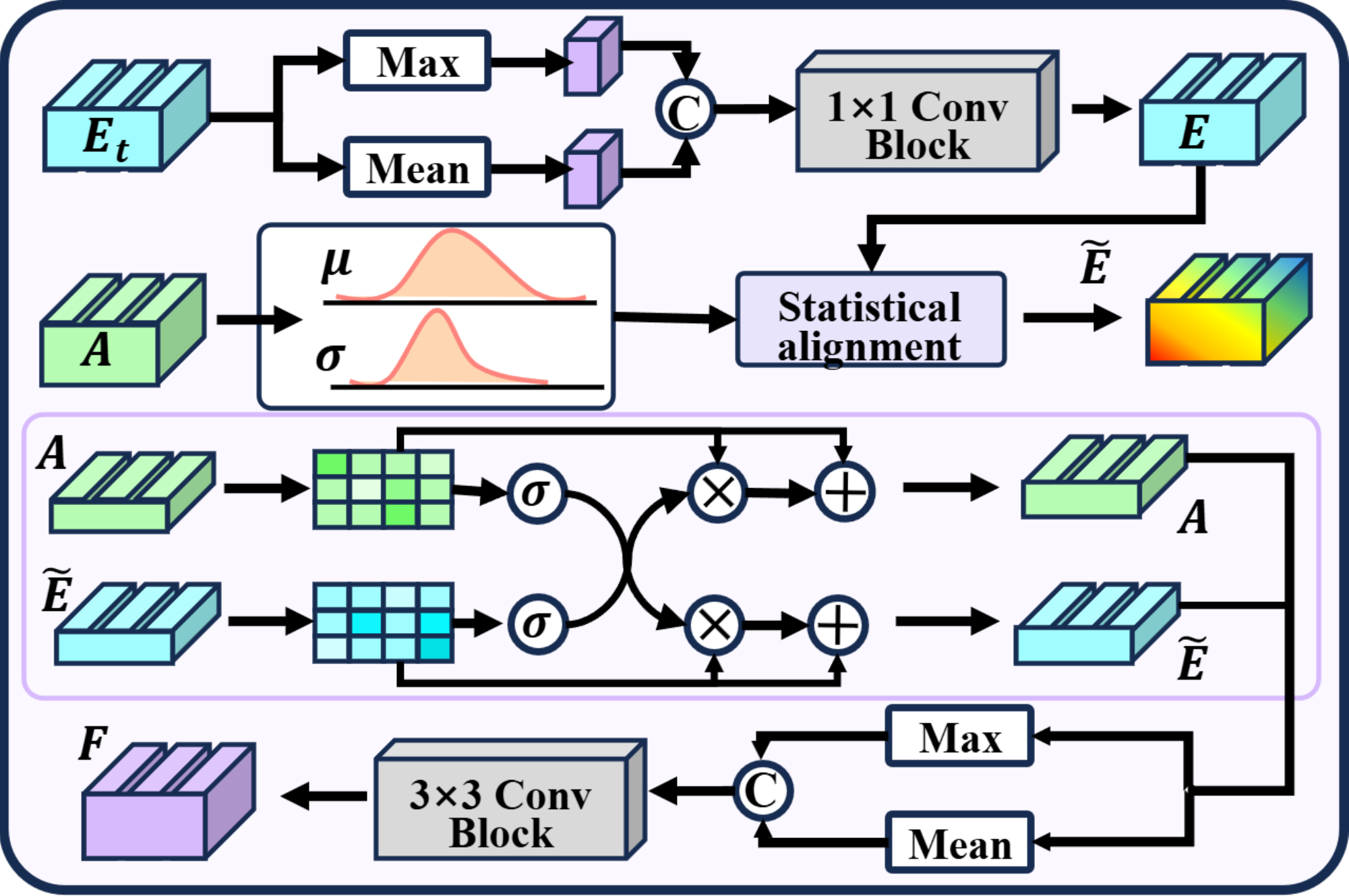}
    \caption{\textbf{Cross-stream interaction module.} The temporal readout pools the SNN responses across the four steps. The ANN features provide the target statistics for alignment. Each branch then generates a gate for the other before fusion.}
\label{fig:cross_stream_interaction}
\end{figure}

\subsection{Cross-Stream Interaction and Measurement Refinement}

\paragraph{Cross-stream interaction.}
As detailed in \cref{fig:cross_stream_interaction}, we first convert each SNN sequence $\bm U_l$ into a dense representation $\bm E_l$ using the max-mean temporal readout (ETA): temporal maximum and average responses are projected separately, then concatenated and mixed by a $1\times1$ convolution block. The maximum preserves strong activations at individual steps, and the mean summarizes the response across the prefix sequence. Both statistics are available to the fusion module. Because ANN and SNN features differ in scale and activation statistics, we next align the SNN readout to the ANN feature using adaptive instance normalization (AdaIN):
\begin{equation}
    \begin{aligned}
        \bm E_l &= \operatorname{ETA} \left(\bm U_l\right),\\
        \widetilde{\bm E}_l
          &=\operatorname{Std} \left(\bm A_l\right)\odot
            \frac{\bm E_l-\operatorname{Mean} \left(\bm E_l\right)}
                 {\operatorname{Std} \left(\bm E_l\right)}
            +\operatorname{Mean} \left(\bm A_l\right),
    \end{aligned}
\end{equation}
where the mean and numerically stabilized standard deviation are computed spatially for each sample and channel. The transformation matches the per-channel spatial mean and standard deviation.

Next, an SREF-style reciprocal gating block allows each representation to modulate the other. Let $\bm g_l^A$ and $\bm g_l^E$ be sigmoid gates obtained from spatial-context pooling and channel transformations of $\bm A_l$ and $\widetilde{\bm E}_l$, respectively. Their residual updates are:
\begin{equation}
    \begin{aligned}
        \bm A_l^+ &= \bm A_l\odot\left(1+\bm g_l^E\right),\\
        \widetilde{\bm E}_l^+
          &=\widetilde{\bm E}_l\odot\left(1+\bm g_l^A\right).
    \end{aligned}
\end{equation}

After branch-specific convolutions, we aggregate the enhanced features using element-wise max and mean across the \textit{two branches}, concatenate both results, and apply a final $3\times3$ convolution block to obtain $\bm F_l$. This block follows the feature interaction in EOLO~\cite{cao2024eolo} and passes the fused features to the measurement-guided correction.

\paragraph{Local count-contrast supervision.}
We derive a structural reference from the counts in the input packet, as illustrated in \cref{fig:spire_overview}(c). Let $\bm N=\sum_{t=1}^{T}\bm S_t$ be the count map and $\bm N_l$ its spatial average pooling at scale $l$. For sensor plane $c$, location $\bm x$, and a one-cell horizontal or vertical offset $\boldsymbol\delta$, define:
\begin{equation}
    m_l^{c,\boldsymbol\delta} \left(\bm x\right)
      =\frac{N_l^c \left(\bm x\right)-N_l^c \left(\bm x+\boldsymbol\delta\right)}
              {N_l^c \left(\bm x\right)+N_l^c \left(\bm x+\boldsymbol\delta\right)+\epsilon}.
\end{equation}

The two directions over four sensor planes yield an eight-channel target $\bm M_l$. Monochrome measurements are repeated across four planes to share the same interface. A binary mask $\boldsymbol\Omega_l$ excludes unsupported regions and invalid neighboring pairs. The denominator normalizes the count difference by the local count level, with $\epsilon>0$ for numerical stability. Each contrast is signed. A positive horizontal contrast, for example, indicates a higher pooled count at the current location than at its horizontal neighbor. The sensor reference preserves both the direction of the change and the plane on which it was measured.

Both branches predict the contrast channels from the ANN feature $\bm A_l$ and the \textit{pre-alignment} SNN readout $\bm E_l$ using independent $1\times1$ convolutions and $\tanh$ activations:
\begin{equation}
    \begin{aligned}
        \widehat{\bm M}_l^a &= \tanh \left(\mathcal C_l^a \left(\bm A_l\right)\right),\\
        \widehat{\bm M}_l^s &= \tanh \left(\mathcal C_l^s \left(\bm E_l\right)\right).
    \end{aligned}
\end{equation}

The shared observation-derived target supervises both predictions through the masked loss
\begin{equation}
    \begin{aligned}
        \mathcal L_{\mathrm{meas}}
          &=\frac{1}{3B}\sum_{l=3}^{5}\sum_{b=1}^{B}\\
          &\quad\times\frac{\sum_{j\in\left\{a,s\right\}}
               \left\lVert \boldsymbol\Omega_{l,b}\odot
               \left(\widehat{\bm M}_{l,b}^j-\bm M_{l,b}\right)\right\rVert_F^2}
                 {\left\lVert\boldsymbol\Omega_{l,b}\right\rVert_1+\epsilon},
    \end{aligned}
\end{equation}
where $b$ indexes the sample, the squared Frobenius norm sums squared errors over contrast channels and spatial locations, and the mask's $\ell_1$ norm counts valid entries.
The dashed region in \cref{fig:spire_overview}(c) denotes the auxiliary supervision used during training.

\paragraph{Bounded measurement-guided correction.}
We project the masked ANN contrast residual into feature channels and modulate it with the SNN readout:
\begin{equation}
    \begin{aligned}
        \bm R_l
          &=\boldsymbol\Omega_l\odot
            \left(\bm M_l-\widehat{\bm M}_l^a\right),\\
        \Delta\bm F_l
          &=\tanh \left(\boldsymbol\alpha_l\right)\odot\rho \left(\bm F_l\right)
          \odot\tanh \left(\mathcal W_l \left(\bm R_l\right)
            \odot\mathcal T_l \left(\bm E_l\right)\right),\\
        \bm F_l'&=\bm F_l+\Delta\bm F_l,
    \end{aligned}
\end{equation}
where $\mathcal W_l$ is a $1\times1$ projection from contrast channels to feature channels, $\mathcal T_l$ transforms the SNN readout, and $\rho \left(\bm F_l\right)$ is the root-mean-square feature magnitude across channels at each location. The channel-wise scale $\boldsymbol\alpha_l$ is initialized to zero, yielding an identity correction at initialization, and the magnitude of each channel update is bounded by the local RMS feature magnitude. At test time, each complete packet yields one set of detections. The forward path uses the observed contrasts, the ANN contrast head, and the pre-alignment SNN readout to produce the correction. The SNN contrast head contributes to the auxiliary training loss.

\paragraph{Detection and optimization.}
Both PANs receive the same corrected pyramid $\left\{\bm F_l'\right\}_{l=3}^{5}$, and their corresponding outputs are averaged before the standard four-level Deformable DETR head~\cite{zhu2021deformable}. The total training objective is
\begin{equation}
    \mathcal L
      =\mathcal L_{\mathrm{det}}+\lambda\mathcal L_{\mathrm{meas}},
\end{equation}
where $\mathcal L_{\mathrm{det}}$ includes Hungarian-matched classification, box regression, generalized IoU, and decoder auxiliary losses. The measurement weight $\lambda$ is ramped up at the start of training.

\section{Experimental Results}

\subsection{Experimental Setup}

\paragraph{Datasets.}
We evaluate SPIRE on PASCAL VOC~\cite{everingham2010voc}, MS COCO~\cite{lin2014coco}, and PKU-Vidar-DVS~\cite{li2022retinomorphic}. For VOC, we train on VOC2007+2012 trainval and evaluate on VOC2007 test across 20 categories, considering underexposure ($0.2\times$) and overexposure ($5.0\times$). For COCO, we use train2017/val2017 with 80 categories. Both datasets are converted into Bayer-patterned spike streams using an integrate-and-fire simulator. PKU-Vidar-DVS contains real monochrome spike recordings across nine categories. We follow its provided training and validation splits and use the monochrome spike recordings as SPIRE's input. The simulated benchmarks test detection at specified exposure conditions, and PKU-Vidar-DVS tests the same method on data from a physical sensor.

\paragraph{Evaluation Metrics.}
We report AP$_{50}$ and AP$_{75}$ at IoU thresholds of 0.50 and 0.75, respectively. Reporting both metrics shows how each method performs when a tighter box match is required.

\paragraph{Implementation Details.}
SPIRE adopts CSPDarknet-Tiny~\cite{wang2020cspnet} and DINOv3-pretrained ConvNeXt-Tiny~\cite{liu2022convnet,simeoni2025dinov3} as semantic backbones. We refer to the two configurations as SPIRE-C and SPIRE-D, respectively. We use 200-frame simulated Bayer spikes at $320\times320$ and 400-frame real Vidar spikes at $416\times416$, with four cumulative SNN steps. Models are trained using AdamW with a learning rate and weight decay of $10^{-4}$, except for DINOv3 parameters with a learning rate of $10^{-5}$. Training lasts 50 epochs on VOC/COCO and 10 epochs on PKU-Vidar-DVS, with a two-epoch warmup, learning rate decay, and EMA evaluation. The measurement loss weight reaches $0.01$ after the ramp-up.

\paragraph{Comparison Methods.}
The simulated VOC and COCO experiments include RGB-event fusion models and detectors that take spike inputs. On PKU-Vidar-DVS, we include methods using real events or real spikes and list the input for each method in the corresponding table.

\subsection{Detection Accuracy}

\begin{table*}[!t]
    \footnotesize
    \centering
    \caption{\textbf{Quantitative comparison on VOC and COCO under exposure degradation.} RGB+E denotes RGB and events. Best and second-best values are shown in \textbf{bold} and \underline{underlined} text, respectively.}
    \label{tab:quantitative}
    \begin{tabularx}{\linewidth}{@{}lccc|CC|CC|CC|CC@{}}
    \toprule[1pt]
    \multirow{3}[4]{*}{Method} & \multirow{3}[4]{*}{Venue} & \multirow{3}[4]{*}{Modality} & \multirow{3}[4]{*}{Params (M)} & \multicolumn{4}{c|}{\textsc{VOC}} & \multicolumn{4}{c}{\textsc{COCO}} \\
    \cmidrule(lr){5-8} \cmidrule(lr){9-12} & & & & \multicolumn{2}{c|}{Underexposure} & \multicolumn{2}{c|}{Overexposure} & \multicolumn{2}{c|}{Underexposure} & \multicolumn{2}{c}{Overexposure} \\
    \cmidrule(lr){5-6} \cmidrule(lr){7-8} \cmidrule(lr){9-10} \cmidrule(lr){11-12} & & & & AP$_{50}$ $\uparrow$ & AP$_{75}$ $\uparrow$ & AP$_{50}$ $\uparrow$ & AP$_{75}$ $\uparrow$ & AP$_{50}$ $\uparrow$ & AP$_{75}$ $\uparrow$ & AP$_{50}$ $\uparrow$ & AP$_{75}$ $\uparrow$ \\
    \midrule
    FPN-Fusion~\cite{FPN-Fusion_icra2022} & ICRA'22 & RGB+E & 65.61 & 56.83 & 29.35 & 52.21 & 26.29 & 31.01 & 14.98 & 29.28 & 13.45 \\
    SODFormer~\cite{li2023sodformer} & TPAMI'23 & RGB+E & 82.03 & 65.37 & 47.39 & 59.65 & 41.95 & 36.19 & 29.78 & 43.17 & 26.74 \\
    RENet~\cite{RENet_icra2024} & ICRA'23 & RGB+E & 373.1 & 57.42 & 31.33 & 52.13 & 27.70 & 33.62 & 16.01 & 31.89 & 14.67 \\
    AFNet~\cite{AFNet_cvpr2023} & CVPR'23 & RGB+E & 26.39 & 55.60 & 28.93 & 51.52 & 26.14 & 32.45 & 15.15 & 30.92 & 13.98 \\
    CAFR~\cite{cao2024cafr} & ECCV'24 & RGB+E & 82.45 & 67.06 & 49.87 & 61.81 & 44.17 & 38.72 & 22.93 & 37.62 & 20.57 \\
    EOLO~\cite{cao2024eolo} & ICRA'24 & RGB+E & 46.29 & 60.61 & 32.70 & 55.87 & 28.97 & 35.24 & 17.59 & 33.10 & 15.21 \\
    SPFD~\cite{wang2026spfd} & CVPR'26 & RGB+E & 78.64 & \underline{75.27} & \underline{57.72} & \underline{72.19} & \underline{55.18} & \underline{47.28} & \underline{30.56} & \underline{44.95} & \underline{28.63} \\
    \midrule
    SpikeYOLO~\cite{luo2024spikeyolo} & ECCV'24 & Spike & 23.16 & 56.43 & 40.49 & 52.31 & 36.75 & 32.69 & 17.29 & 31.37 & 15.96 \\
    CREST~\cite{mao2025crest} & AAAI'25 & Spike & 7.66 & 23.22 & 9.73 & 19.27 & 5.86 & 19.86 & 8.62 & 17.92 & 7.16 \\
    TDE~\cite{luo2026tde} & AAAI'26 & Spike & 23.32 & 35.27 & 23.24 & 33.17 & 20.75 & 28.96 & 16.37 & 26.15 & 14.63 \\
    \rowcolor{gray!20}
    SPIRE-C (Ours) & - & Spike & 44.04 & 63.98 & 47.77 & 60.24 & 44.71 & 41.26 & 26.61 & 39.68 & 25.17 \\
    \rowcolor{gray!20} 
    SPIRE-D (Ours) & - & Spike & 69.89 & \textbf{79.59} & \textbf{63.23} & \textbf{77.69} & \textbf{60.32} & \textbf{51.28} & \textbf{35.29} & \textbf{49.48} & \textbf{32.97} \\
    \bottomrule[1pt]
    \end{tabularx}%
\end{table*}

\paragraph{Simulated under- and overexposure.}
SPIRE-D leads on both AP metrics across the four settings in \cref{tab:quantitative}. On VOC, it reaches 79.59/63.23 AP$_{50}$/AP$_{75}$ under underexposure and 77.69/60.32 under overexposure. Relative to SPFD, the strongest baseline in the table, the gains are 4.32/5.51 and 5.50/5.14 points, respectively. On COCO, the AP$_{50}$ gains are 4.00 and 4.53 points, and the corresponding AP$_{75}$ gains are 4.73 and 4.34 points. The gains thus extend to detection at the stricter overlap threshold on both datasets.

Across the two tested exposure settings, SPIRE-D's AP$_{50}$ decreases by 1.90 points on VOC and 1.80 on COCO from underexposure to overexposure. The corresponding decreases for SPFD are 3.08 and 2.33 points. SPIRE-C also exceeds SpikeYOLO, the highest-scoring spike-input baseline in the simulated comparison, by 7.55-8.57 AP$_{50}$ points across these four settings. This comparison shows that the compact configuration also offers a consistent gain within the spike input modality.

\begin{table}[!t]
    \footnotesize
    \centering
    \caption{\textbf{Quantitative comparison on PKU-Vidar-DVS.} Best and second-best are shown in \textbf{bold} and \underline{underlined}, respectively.}
    \label{tab:pku_comparison}
    \setlength{\tabcolsep}{3pt}
    \begin{tabularx}{\linewidth}{@{}llc|CC@{}}
        \toprule[1pt]
        Method & Venue & Input  & AP$_{50}$ $\uparrow$ & AP$_{75}$ $\uparrow$ \\
        \midrule
        E2VID+YOLO~\cite{rebecq2019eventstovideo} & CVPR'19 & Event & 39.40 & - \\
        TAR-Events+YOLO~\cite{li2022retinomorphic} & AAAI'22 & Event & 38.60 & - \\
        CREST~\cite{mao2025crest} & AAAI'25 & Event & 43.20 & - \\
        SSLA-Det~\cite{hao2026ssla} & ECCV'26 & Event & 31.86 & 8.45 \\
        \midrule        
        TAR-Spikes+YOLO~\cite{li2022retinomorphic} & AAAI'22 & Spike & 57.90 & - \\
        VTII+YOLO~\cite{li2022retinomorphic} & AAAI'22 & Spike & 55.10 & - \\
        VTTW+YOLO~\cite{li2022retinomorphic} & AAAI'22 & Spike & 51.60 & - \\
        SpikeYOLO~\cite{luo2024spikeyolo} & ECCV'24 & Spike & 62.98 & 38.64 \\
        VTTI+RT-DETR~\cite{zhao2024rtdetr,liu2025easd} & CVPR'24 & Spike & 46.00 & - \\
        EASD~\cite{liu2025easd} & arXiv'25 & Spike & 60.10 & - \\
        CREST~\cite{mao2025crest} & AAAI'25 & Spike & 29.13 & 11.80 \\
        SPFD~\cite{wang2026spfd} & CVPR'26 & Spike & 63.12 & \underline{39.58} \\
        TDE~\cite{luo2026tde} & AAAI'26 & Spike & 49.18 & 25.30 \\
        \rowcolor{gray!20}
        SPIRE-C (Ours) & - & Spike & \underline{63.27} & 37.76 \\
        \rowcolor{gray!20}
        SPIRE-D (Ours) & - & Spike & \textbf{67.07} & \textbf{40.10} \\
        \bottomrule[1pt]
    \end{tabularx}
\end{table}

\paragraph{Real spike recordings.}
On PKU-Vidar-DVS (\cref{tab:pku_comparison}), SPIRE-D reaches 67.07 AP$_{50}$ and 40.10 AP$_{75}$, exceeding SPFD by 3.95 and 0.52 points, respectively. The larger gain at IoU 0.50 shows that the advantage is stronger at the more permissive threshold on these real recordings. SPIRE-C reaches 63.27 AP$_{50}$, close to SPFD's 63.12, while its AP$_{75}$ is 1.82 points lower. Among the other methods with spike inputs, SPIRE-D improves AP$_{50}$ by 4.09 points over SpikeYOLO and by 6.97 points over EASD.

\subsection{Computational Efficiency}

\begin{table}[t]
    \centering
    \footnotesize
    \setlength{\tabcolsep}{2pt}
    \caption{\textbf{Detection accuracy and inference cost on PKU-Vidar-DVS and VOC-UE.} PKU denotes PKU-Vidar-DVS, and VOC-UE denotes underexposed VOC. Best and second-best values are shown in \textbf{bold} and \underline{underlined} text, respectively.}
    \begin{tabularx}{\linewidth}{@{}l|ccc|CC|CC@{}}
    \toprule[1pt]
    \multirow{2}[2]{*}{Method} & \multirow{2}[2]{*}{\makecell{Params\\(M) }} & \multirow{2}[2]{*}{{MACs}} & \multirow{2}[2]{*}{\makecell{Time\\(ms) }} & \multicolumn{2}{c|}{PKU} & \multicolumn{2}{c}{VOC-UE} \\
    \cmidrule(lr){5-6}\cmidrule(lr){7-8}
    & & & & AP$_{50}$  & AP$_{75}$  & AP$_{50}$  & AP$_{75}$  \\
    \midrule
    SpikeYOLO~\cite{luo2024spikeyolo} & \textbf{23.126} & \underline{51.2} & \textbf{36.3} & 62.98 & 38.64 & 56.43 & 40.49 \\
    SPFD~\cite{wang2026spfd} & 78.637 & 126.3 & 65.4 & 63.12 & \underline{39.58} & \underline{75.27} & \underline{57.72} \\
    \rowcolor{gray!20}
    {SPIRE-C} & \underline{44.044} & \textbf{46.7} & \underline{56.8} & \underline{63.27} & 37.76 & 63.98 & 47.77 \\
    \rowcolor{gray!20} {SPIRE-D} & 69.887 & 62.7 & 60.9 & \textbf{67.07} & \textbf{40.10} & \textbf{79.59} & \textbf{63.23} \\
    \bottomrule[1pt]
    \end{tabularx}
    \label{tab:pku_efficiency}
\end{table}

\cref{tab:pku_efficiency} compares accuracy with parameter count, MACs, and inference latency. SPIRE-D uses 50.4\% fewer MACs than SPFD (62.7 vs. 126.3), with a latency of 60.9 ms versus 65.4 ms and higher AP at both IoU thresholds on the two benchmarks. SPIRE-C has the lowest MAC count (46.7). Compared with SpikeYOLO, SPIRE-C gains 7.55 AP$_{50}$ points on underexposed VOC and 0.29 points on PKU-Vidar-DVS, with 20.5 ms higher measured latency.

Within SPIRE, moving from the compact to the larger configuration adds 4.1 ms of latency and improves AP$_{50}$ by 3.80 points on PKU-Vidar-DVS and 15.61 points on underexposed VOC. The accuracy return from the larger configuration is thus substantially greater on underexposed VOC in this comparison.

\subsection{Ablation Studies}

All ablations use SPIRE-C to keep the semantic backbone fixed. This minimizes the influence of a powerful semantic backbone, allowing us to better isolate the contributions of the proposed modules. We train for 10 epochs on PKU-Vidar-DVS and 50 epochs on VOC under both exposures.

\begin{table}[!t]
    \footnotesize
    \centering
    \caption{\textbf{Ablation of key components in SPIRE-C.} UE and OE denote under- and overexposure, respectively. Best and second-best values are shown in \textbf{bold} and \underline{underlined}, respectively.}
    \label{tab:component_ablation}
    \setlength{\tabcolsep}{3pt}
    \begin{tabularx}{\linewidth}{@{}l|CC|CC@{}}
        \toprule[1pt]
        \multirow{2}[2]{*}{Variant}
        & \multicolumn{2}{c|}{PKU-Vidar-DVS} & \multicolumn{2}{c}{\textsc{VOC}} \\
        \cmidrule(lr){2-3} \cmidrule(lr){4-5} & AP$_{50}$ $\uparrow$ & AP$_{75}$ $\uparrow$ & AP$_{50}^{\mathrm{UE}}$ $\uparrow$ & AP$_{50}^{\mathrm{OE}}$ $\uparrow$ \\
        \midrule
        ANN-only & 59.26 & 32.23 & 60.50 & 56.36 \\
        \textit{w/o} ISP & 61.00 & 35.28 & 56.52 & 53.40 \\
        \textit{w/o} Cumulative Views & 61.27 & 35.71 & 59.62 & 56.97 \\
        \textit{w/o} Semantic Prompt & \underline{62.44} & \underline{35.98} & 61.37 & 57.29 \\
        \textit{w/o} Cross-stream Interaction & 59.13 & 32.17 & \underline{61.59} & \underline{57.83} \\
        \textit{w/o} Measure. Refine. & 60.71 & 34.37 & 60.51 & 56.52 \\
        \textit{w/o} Dual PAN & 60.39 & 33.97 & 60.37 & 57.69 \\
        \rowcolor{gray!20}
        Full Model & \textbf{63.27} & \textbf{37.76} & \textbf{63.98} & \textbf{60.24} \\
        \bottomrule[1pt]
    \end{tabularx}
\end{table}

\paragraph{Contribution of each component.}
In \cref{tab:component_ablation}, adding the spiking branch and associated modules to the ANN-only configuration improves AP$_{50}$ by 4.01 points on PKU-Vidar-DVS and by 3.48 and 3.88 points under VOC under- and overexposure. The largest single-removal effects differ by setting. Removing the adaptive ISP gives the largest AP$_{50}$ drops among the component removals on VOC (7.46 and 6.84 points), consistent with its role in adapting the appearance input to exposure. On real PKU-Vidar-DVS, removing cross-stream interaction gives the largest drop (4.14 points), suggesting that how the two feature streams are combined matters particularly on this benchmark. Removing measurement-guided refinement lowers AP$_{50}$ in all three settings, by 2.56 points on PKU-Vidar-DVS and by 3.47 and 3.72 points on VOC.

\begin{table}[!t]
    \footnotesize
    \centering
    \caption{\textbf{Ablation of temporal inputs and neuronal dynamics.} Static denotes a full-window input, and cumulative denotes the four ordered prefix views. Best and second-best values are shown in \textbf{bold} and \underline{underlined} text, respectively.}
    \label{tab:temporal_ablation}
    \setlength{\tabcolsep}{3pt}
    \begin{tabularx}{\linewidth}{@{}lc|CC|CC@{}}
        \toprule[1pt]
        \multirow{2}[2]{*}{Input} & \multirow{2}[2]{*}{Neuron} & \multicolumn{2}{c|}{PKU-Vidar-DVS} & \multicolumn{2}{c}{\textsc{VOC}} \\
        \cmidrule(lr){3-4} \cmidrule(lr){5-6} & & AP$_{50}$ $\uparrow$ & AP$_{75}$ $\uparrow$ & AP$_{50}^{\mathrm{UE}}$ $\uparrow$ & AP$_{50}^{\mathrm{OE}}$ $\uparrow$ \\
        \midrule
        Static & Non-spiking & \underline{61.76} & 35.12 & \underline{64.08} & \underline{60.49} \\
        Static & LIF & 61.27 & \underline{35.71} & 59.62 & 56.97 \\
        Cumulative & Non-spiking & 61.22 & 35.36 & \textbf{64.48} & \textbf{60.61} \\
        \rowcolor{gray!20}
        Cumulative & LIF & \textbf{63.27} & \textbf{37.76} & 63.98 & 60.24 \\
        \bottomrule[1pt]
    \end{tabularx}
\end{table}

\paragraph{Cumulative inputs and neuronal dynamics.}
\cref{tab:temporal_ablation} varies the input representation and neuron type. Under LIF encoding, cumulative inputs improve AP$_{50}$ by 2.00 points on PKU-Vidar-DVS and by 4.36 and 3.27 points under VOC under- and overexposure. With cumulative inputs fixed, LIF improves AP$_{50}$/AP$_{75}$ by 2.05/2.40 points over non-spiking encoding on real recordings. On VOC, the cumulative non-spiking variant instead leads the cumulative LIF variant by 0.50 and 0.37 AP$_{50}$ points. On the same real data, the static LIF variant scores 0.49 AP$_{50}$ points below the static non-spiking variant. The advantage of LIF on PKU-Vidar-DVS thus depends on giving the encoder a sequence of cumulative inputs in this ablation.

\begin{table}[!t]
    \footnotesize
    \centering
    \caption{\textbf{Ablation of cumulative inputs and semantic-guided neuron configuration.} Cum. denotes cumulative inputs. Best and second-best are shown in \textbf{bold} and \underline{underlined}, respectively.}
    \label{tab:semantic_ablation}
    \begin{tabularx}{\linewidth}{@{}cc|CC|CC@{}}
        \toprule[1pt]
        \multirow{2}[2]{*}{Cum.} & \multirow{2}[2]{*}{Prompt} & \multicolumn{2}{c|}{PKU-Vidar-DVS} & \multicolumn{2}{c}{\textsc{VOC}} \\
        \cmidrule(lr){3-4} \cmidrule(lr){5-6} & & AP$_{50}$ $\uparrow$ & AP$_{75}$ $\uparrow$ & AP$_{50}^{\mathrm{UE}}$ $\uparrow$ & AP$_{50}^{\mathrm{OE}}$ $\uparrow$ \\
        \midrule
        \xmark & \xmark & 60.19 & 33.58 & 58.81 & 55.27 \\
        \cmark & \xmark & \underline{62.44} & \underline{35.98} & \underline{61.37} & \underline{57.29} \\
        \xmark & \cmark & 61.27 & 35.71 & 59.62 & 56.97 \\
        \rowcolor{gray!20}
        \cmark & \cmark & \textbf{63.27} & \textbf{37.76} & \textbf{63.98} & \textbf{60.24} \\
        \bottomrule[1pt]
    \end{tabularx}
\end{table}

\paragraph{Semantic prompts during spike encoding.}
\cref{tab:semantic_ablation} crosses the use of cumulative inputs with the use of semantic prompts. With cumulative inputs fixed, prompting improves AP$_{50}$/AP$_{75}$ by 0.83/1.78 points on PKU-Vidar-DVS. On VOC, the AP$_{50}$ gains from prompting are 2.61 and 2.95 points with cumulative inputs, compared with 0.81 and 1.70 points with static inputs. On VOC, cumulative inputs increase the gain from prompting by 1.80 points under underexposure and 1.25 points under overexposure. On PKU-Vidar-DVS, prompting improves AP$_{75}$ by 2.13 points with static inputs and 1.78 points with cumulative inputs. The prompt improves detection with either input on real recordings, while the larger prompt gains with cumulative inputs are observed for VOC AP$_{50}$.

\begin{table}[!t]
    \footnotesize
    \centering
    \caption{\textbf{Ablation of measurement-guided refinement.} Supervision enables the auxiliary count-contrast loss during training. Correction enables the masked residual update to the fused pyramid. Best and second-best values are shown in \textbf{bold} and \underline{underlined} text, respectively.}
    \label{tab:measurement_ablation}
    \setlength{\tabcolsep}{3pt}
    \begin{tabularx}{\linewidth}{@{}cc|CC|CC@{}}
        \toprule[1pt]
        \multirow{2}[2]{*}{Supervision} & \multirow{2}[2]{*}{Correction} & \multicolumn{2}{c|}{PKU-Vidar-DVS} & \multicolumn{2}{c}{\textsc{VOC}} \\
        \cmidrule(lr){3-4} \cmidrule(lr){5-6} & & AP$_{50}$ $\uparrow$ & AP$_{75}$ $\uparrow$ & AP$_{50}^{\mathrm{UE}}$ $\uparrow$ & AP$_{50}^{\mathrm{OE}}$ $\uparrow$ \\
        \midrule
        \xmark & \xmark & 60.71 & 34.37 & \underline{60.51} & \underline{56.52} \\
        \cmark & \xmark & 60.94 & 34.26 & 58.66 & 55.05 \\
        \xmark & \cmark & \underline{61.31} & \underline{35.05} & 59.90 & 54.96 \\
        \rowcolor{gray!20}
        \cmark & \cmark & \textbf{63.27} & \textbf{37.76} & \textbf{63.98} & \textbf{60.24} \\
        \bottomrule[1pt]
    \end{tabularx}
\end{table}

\begin{figure*}[!t]
    \centering
    \setlength{\tabcolsep}{1pt}
    \begin{tabular}{@{}cccccc@{}}
     Spike frame & SPFD~\cite{wang2026spfd} & EOLO~\cite{cao2024eolo} & \textbf{SPIRE-D (Ours)} & Heatmap (Ours) & Ground truth \\
    \includegraphics[width=0.1605\linewidth]{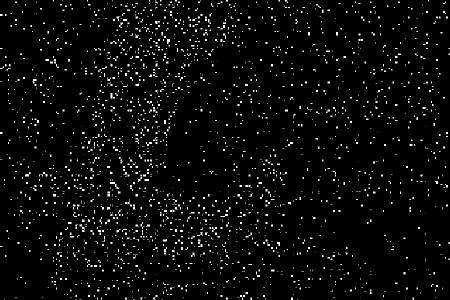} & \includegraphics[width=0.1605\linewidth]{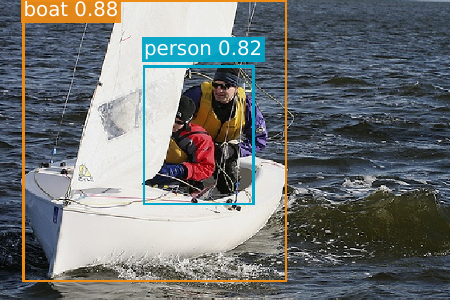} & \includegraphics[width=0.1605\linewidth]{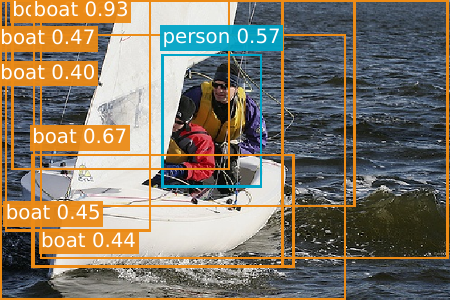} & \includegraphics[width=0.1605\linewidth]{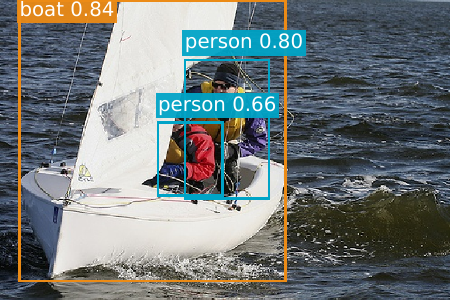} & \includegraphics[width=0.1605\linewidth]{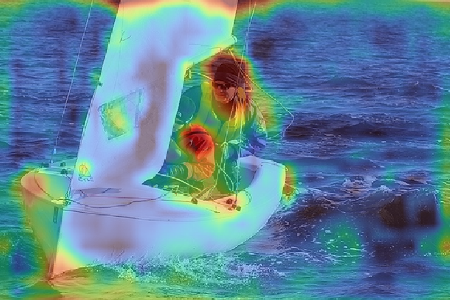} & \includegraphics[width=0.1605\linewidth]{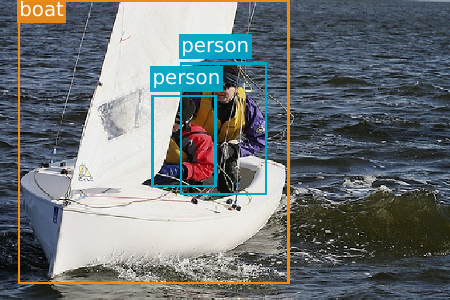} \\ 
    \includegraphics[width=0.1605\linewidth]{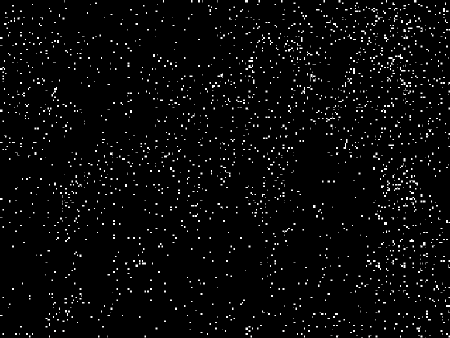} & \includegraphics[width=0.1605\linewidth]{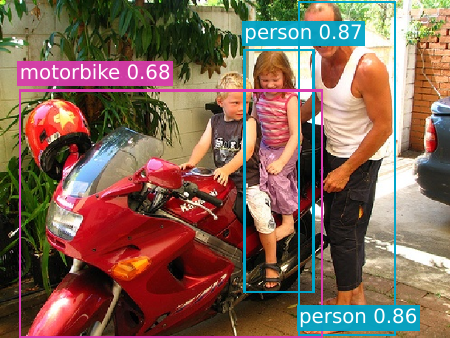} & \includegraphics[width=0.1605\linewidth]{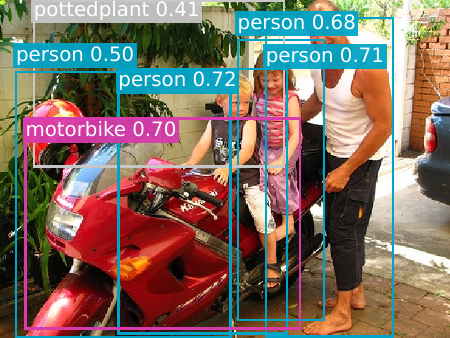} & \includegraphics[width=0.1605\linewidth]{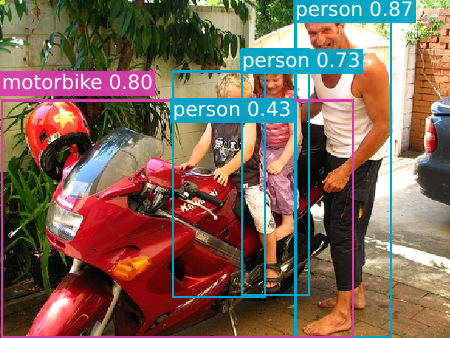} & \includegraphics[width=0.1605\linewidth]{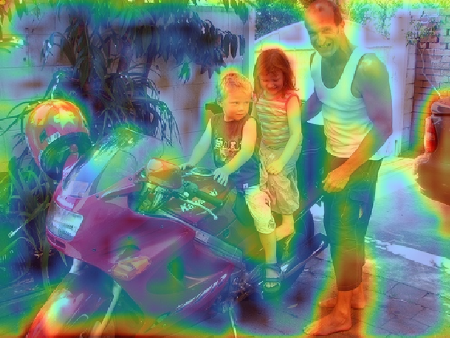} & \includegraphics[width=0.1605\linewidth]{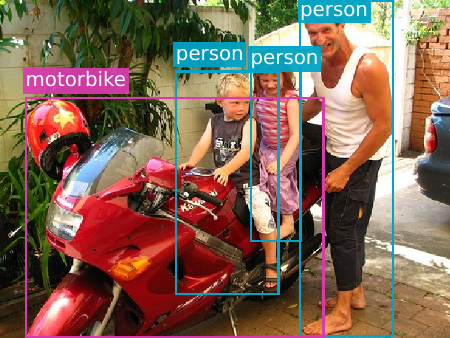} \\ 
    \includegraphics[width=0.1605\linewidth]{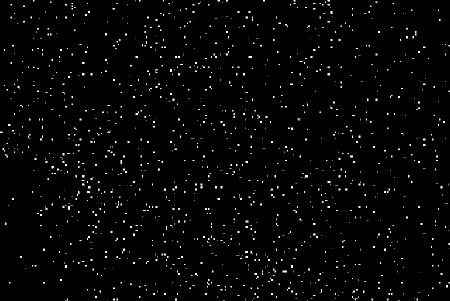} & \includegraphics[width=0.1605\linewidth]{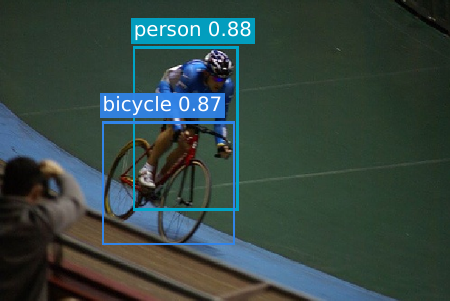} & \includegraphics[width=0.1605\linewidth]{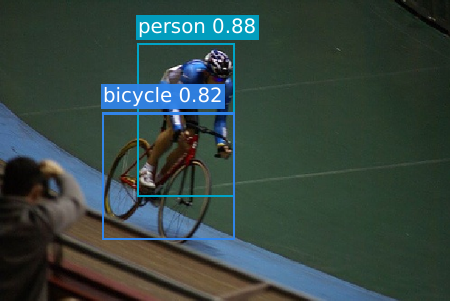} & \includegraphics[width=0.1605\linewidth]{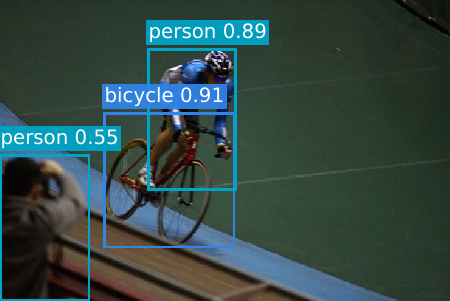} & \includegraphics[width=0.1605\linewidth]{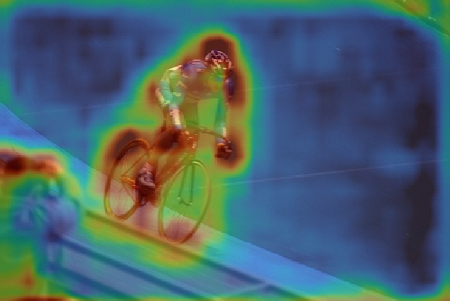} & \includegraphics[width=0.1605\linewidth]{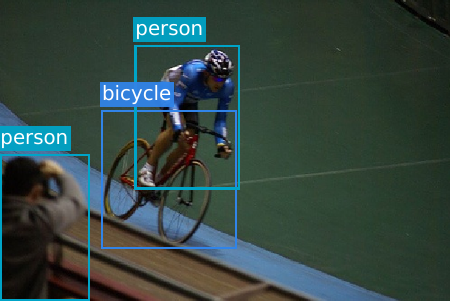} \\ 
    \includegraphics[width=0.1605\linewidth]{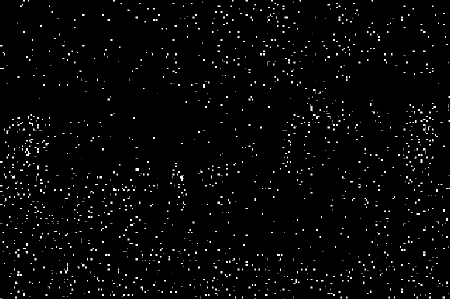} & \includegraphics[width=0.1605\linewidth]{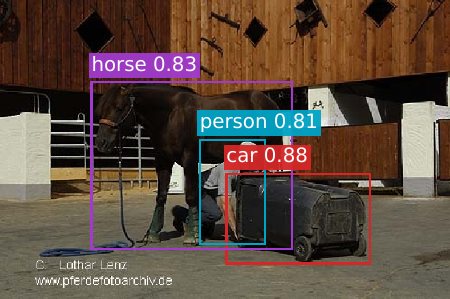} & \includegraphics[width=0.1605\linewidth]{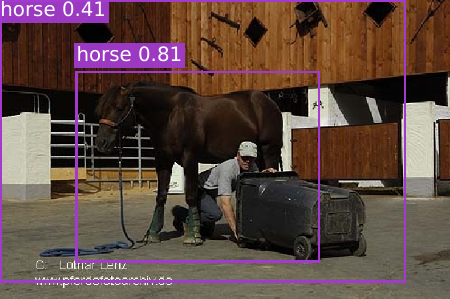} & \includegraphics[width=0.1605\linewidth]{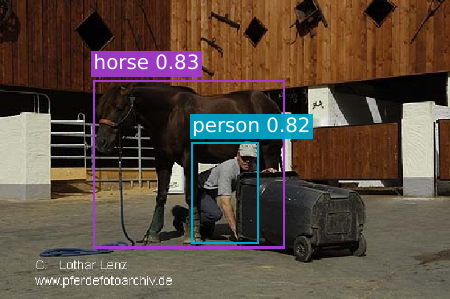} & \includegraphics[width=0.1605\linewidth]{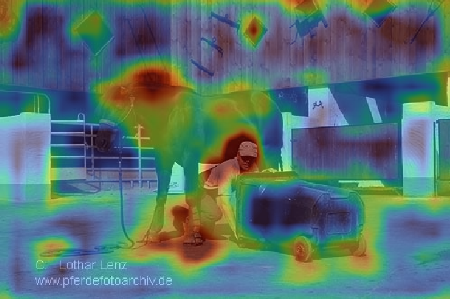} & \includegraphics[width=0.1605\linewidth]{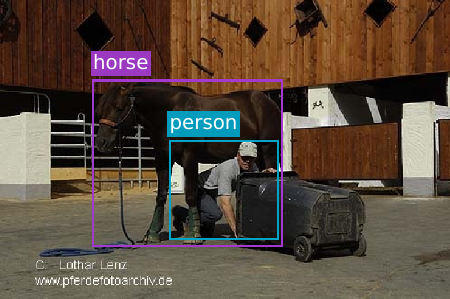} \\

    \end{tabular}
    \caption{\textbf{Qualitative comparisons on VOC underexposure scenes.} The columns show the spike frames, SPFD, EOLO, SPIRE-D, SPIRE-D feature heatmaps, and ground-truth boxes.}
    \label{fig:spire_voc_ue_qualitative_comparison}
\end{figure*}

\paragraph{Measurement supervision and residual correction.}
\cref{tab:measurement_ablation} isolates the training-time contrast supervision from the feature correction. On PKU-Vidar-DVS, each component alone changes AP$_{50}$ by at most 0.60 points, while enabling both improves AP$_{50}$/AP$_{75}$ by 2.56/3.39 points over the variant with neither. On VOC, either component alone lowers AP$_{50}$ under both exposure settings. Enabling both raises AP$_{50}$ by 3.47 and 3.72 points over the variant with neither. The effect of enabling correction depends on whether the contrast predictions are supervised. With supervision, correction raises AP$_{50}$ by 2.33 points on PKU-Vidar-DVS and by 5.32 and 5.19 points under VOC under- and overexposure. Without supervision, the change is +0.60, -0.61, and -1.56 points, respectively. This pattern supports the role of the supervised contrast predictions in making the residual correction useful for detection.

\subsection{Qualitative Results and Class-Wise Recall}

\paragraph{Detection examples.}
\cref{fig:spire_voc_ue_qualitative_comparison} shows representative underexposed VOC scenes. In these examples, SPIRE-D produces fewer redundant boxes and closer matches to the annotated boxes than SPFD and EOLO. The accompanying heatmaps show the spatial distribution of SPIRE-D's feature responses around the objects. Notably, SPIRE-D maintains reliable detections in scenes containing multiple objects and weakened visual details, where competing methods exhibit missed or inaccurate predictions. These results demonstrate the effectiveness of SPIRE in maintaining robust object perception from sparse spike observations.

\begin{figure}[!t]
    \centering
    \setlength{\tabcolsep}{2pt}
    \begin{tabular}{@{}cc@{}}
    \includegraphics[width=0.49\linewidth]{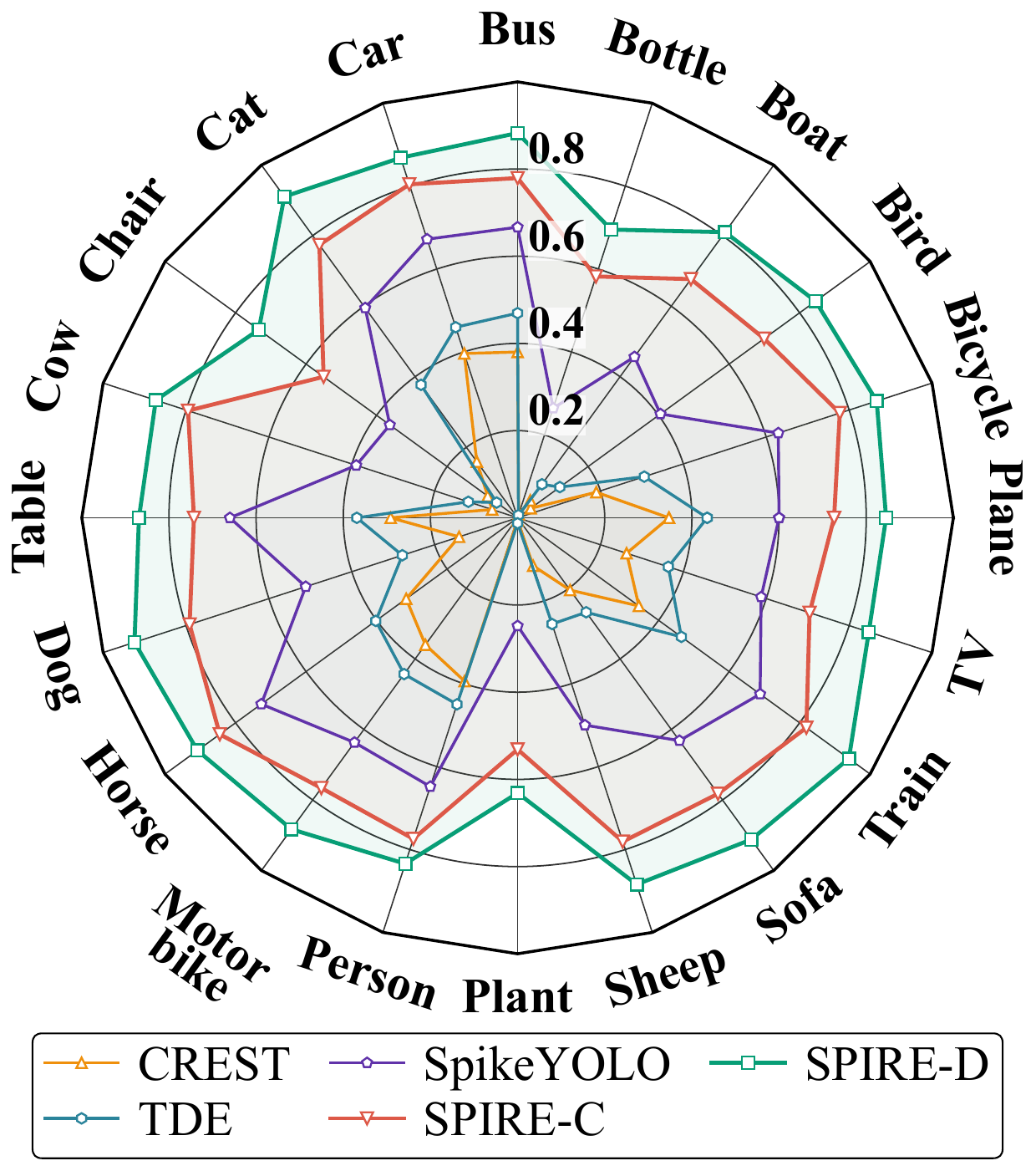} &
    \includegraphics[width=0.49\linewidth]{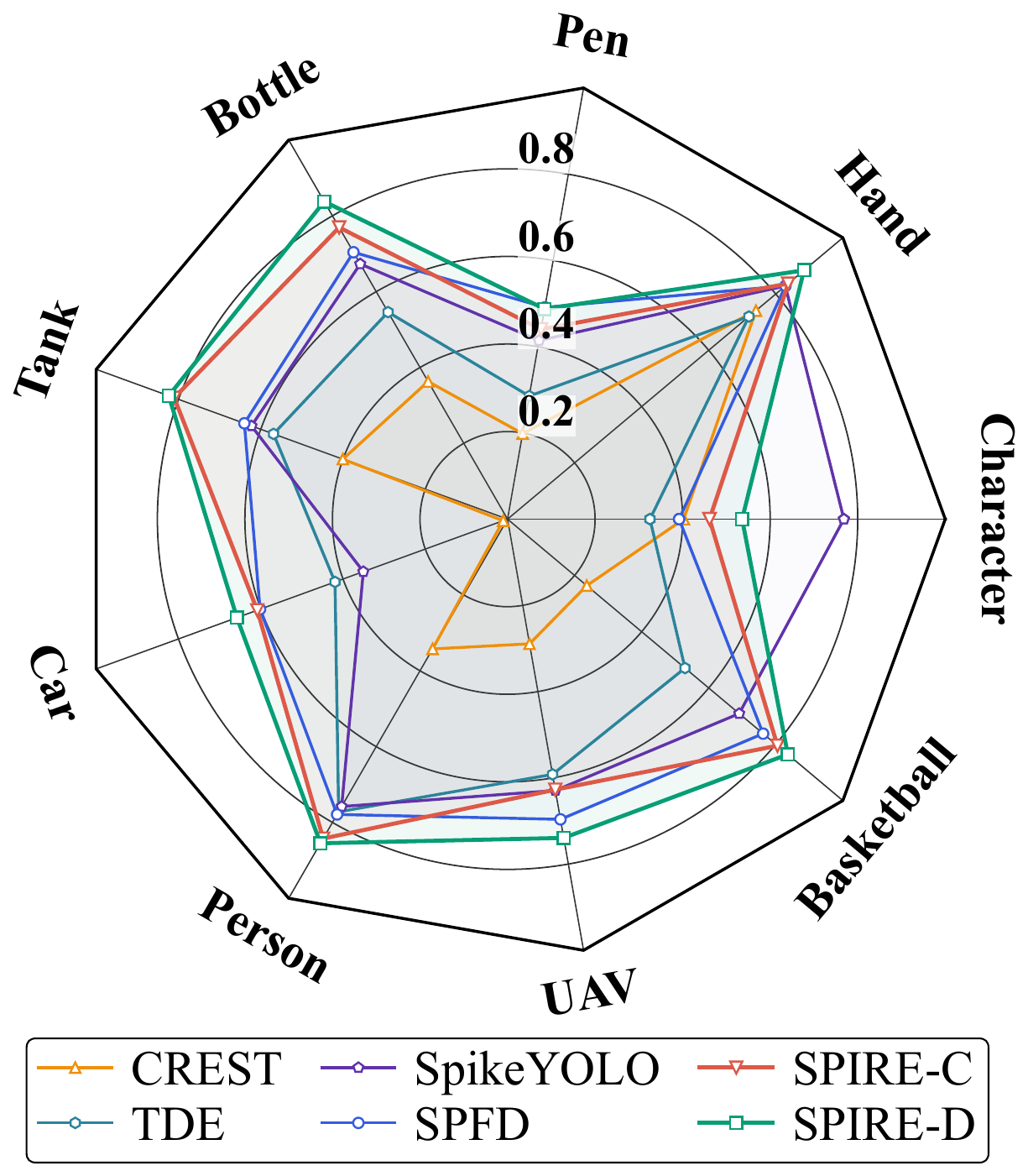} \\
    \footnotesize (a) VOC UE & \footnotesize (b) PKU-Vidar-DVS \\
    \end{tabular}
    \caption{\textbf{Class-wise recall under simulated and real spike scenes.} (a) VOC underexposure. (b) PKU-Vidar-DVS.}
    \label{fig:recall_radar_comparison}
\end{figure}

\paragraph{Class-wise recall.}
\cref{fig:recall_radar_comparison} compares class-wise recall on underexposed VOC and real PKU-Vidar-DVS. SPIRE-D achieves higher recall across multiple categories, while SPIRE-C remains competitive despite its lightweight backbone. The improvements are observed in both simulated low-light conditions and real spike recordings, indicating the robustness of SPIRE across different observation settings. These results suggest that complementary semantic and temporal-structural representations help preserve discriminative object evidence, reducing missed detections under challenging illumination and real spike observations.

\section{Conclusion}

We propose SPIRE, a sensor-grounded semantic-spiking illumination-robust estimation for object detection using a single spike modality. SPIRE constructs complementary semantic and temporal-structural representations, with semantic cues configuring membrane potentials and firing thresholds before cross-stream fusion. Measurement-guided refinement further incorporates raw spike-count evidence into feature correction. Extensive experiments on simulated and real spike recordings demonstrate robust detection performance with favorable accuracy-efficiency trade-offs.

\section*{LLM Usage Statement}

We used ChatGPT and Codex (OpenAI) for language editing of the
manuscript and supplementary material. We also used these tools
for notation checks and LaTeX formatting.

{
    \small
    \bibliographystyle{ieeenat_fullname}
    \bibliography{SPIRE}
}

\end{document}